\documentclass{article}
\usepackage{icmc2024template}
\usepackage{times}
\usepackage{ifpdf}
\usepackage{soul}
\usepackage{amsmath}
\usepackage[english]{babel}
\usepackage{graphicx}
\usepackage{subcaption}

\def\papertitle{Long-Term, Store-Front Robotics: Interactive Music for Robotic Arm, Caxixi and Frame Drums}
\def\firstauthor{Richard Savery}
\def\secondauthor{Fouad Sukkar}
\def\thirdauthor{Third Author}

\newif\ifpdf
\ifx\pdfoutput\relax
\else
   \ifcase\pdfoutput
      \pdffalse
   \else
      \pdftrue
  \fi
\fi

\ifpdf 
  \usepackage[pdftex,
    pdftitle={\papertitle},
    pdfauthor={\firstauthor, \secondauthor, \thirdauthor},
    bookmarksnumbered, 
    pdfstartview=XYZ 
   ]{hyperref}

  \usepackage[figure,table]{hypcap}

\else 
  \usepackage[dvips,
    bookmarksnumbered, 
    pdfstartview=XYZ 
  ]{hyperref}  

  \usepackage[dvips]{epsfig,graphicx}
  \graphicspath{{./figures/}}
  \DeclareGraphicsExtensions{.eps}

  \usepackage[figure,table]{hypcap}
\fi

\hypersetup{
    colorlinks,%
    citecolor=black,%
    filecolor=black,%
    linkcolor=black,%
    urlcolor=black
}

\title{\papertitle}

\twoauthors
  {1.5in}
  {\firstauthor} {Macquarie University \\  %
    {\tt \href{mailto:richard.savery@mq.edu.au}{richard.savery@mq.edu.au}}}
  {\secondauthor} {University of Technology Sydney \\  %
                    Australian Cobotics Centre \\ %
    {\tt \href{mailto:fouad.sukkar@uts.edu.au}{fouad.sukkar@uts.edu.au}}}

\begin{document}
\capstartfalse
\maketitle
\capstarttrue
\begin{abstract}
This paper presents an innovative exploration into the integration of interactive robotic musicianship within a commercial retail environment, specifically through a three-week-long in-store installation featuring a UR3 robotic arm, custom-built frame drums, and an adaptive music generation system. Situated in a prominent storefront in one of the world's largest cities, this project aimed to enhance the shopping experience by creating dynamic, engaging musical interactions that respond to the store's ambient soundscape. Key contributions include the novel application of industrial robotics in artistic expression, the deployment of interactive music to enrich retail ambiance, and the demonstration of continuous robotic operation in a public setting over an extended period. Challenges such as system reliability, variation in musical output, safety in interactive contexts, and brand alignment were addressed to ensure the installation's success. The project not only showcased the technical feasibility and artistic potential of robotic musicianship in retail spaces but also offered insights into the practical implications of such integration, including system reliability, the dynamics of human-robot interaction, and the impact on store operations. This exploration opens new avenues for enhancing consumer retail experiences through the intersection of technology, music, and interactive art, suggesting a future where robotic musicianship contributes meaningfully to public and commercial spaces.

\end{abstract}

 \begin{figure}[t]
	\centering
		\includegraphics[width=\columnwidth]{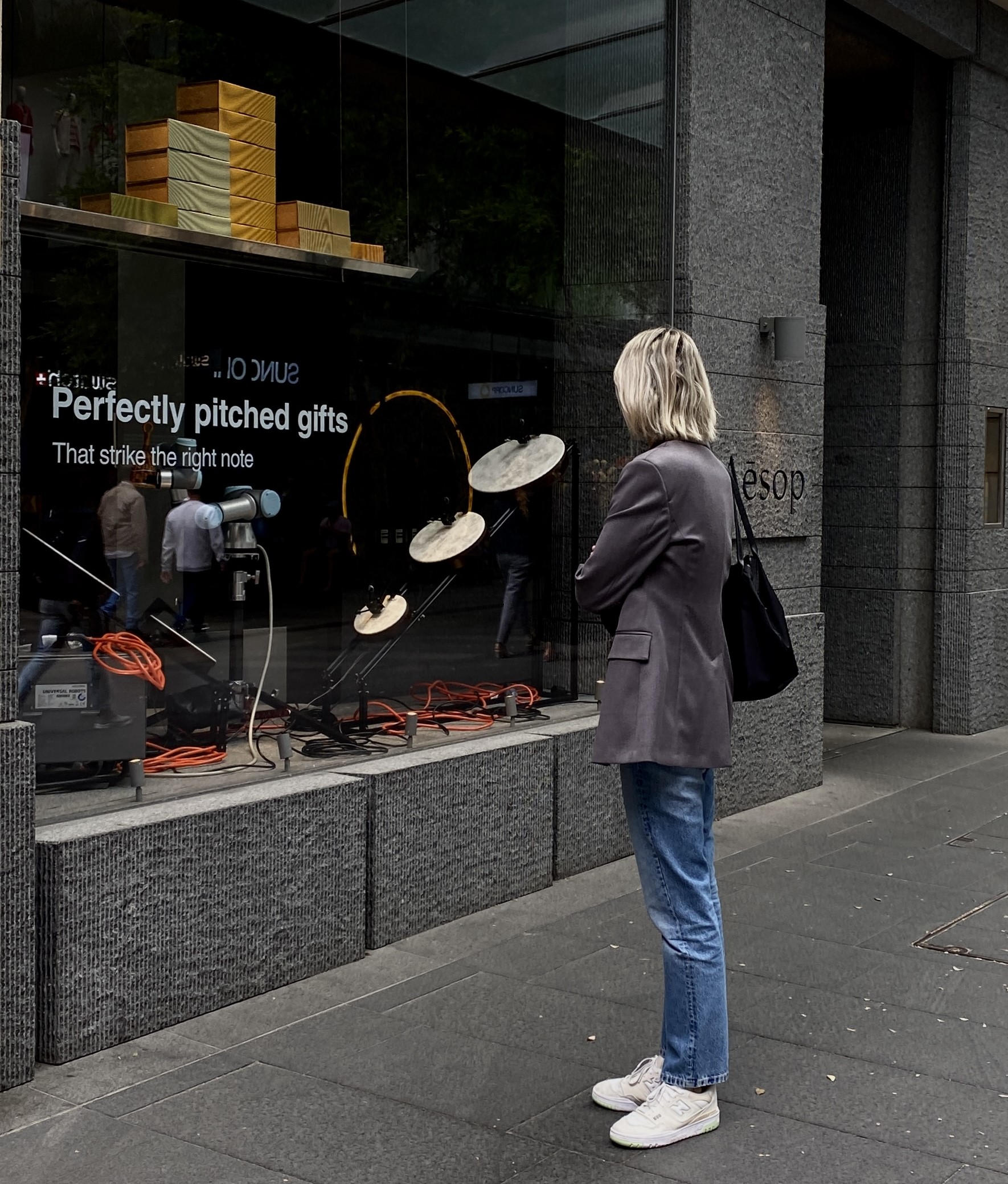}
	\caption{Photo of store-front display}
	\label{fig:current.png}
\end{figure}

\begin{figure*}[t]
	\centering
		\includegraphics[width=2\columnwidth]{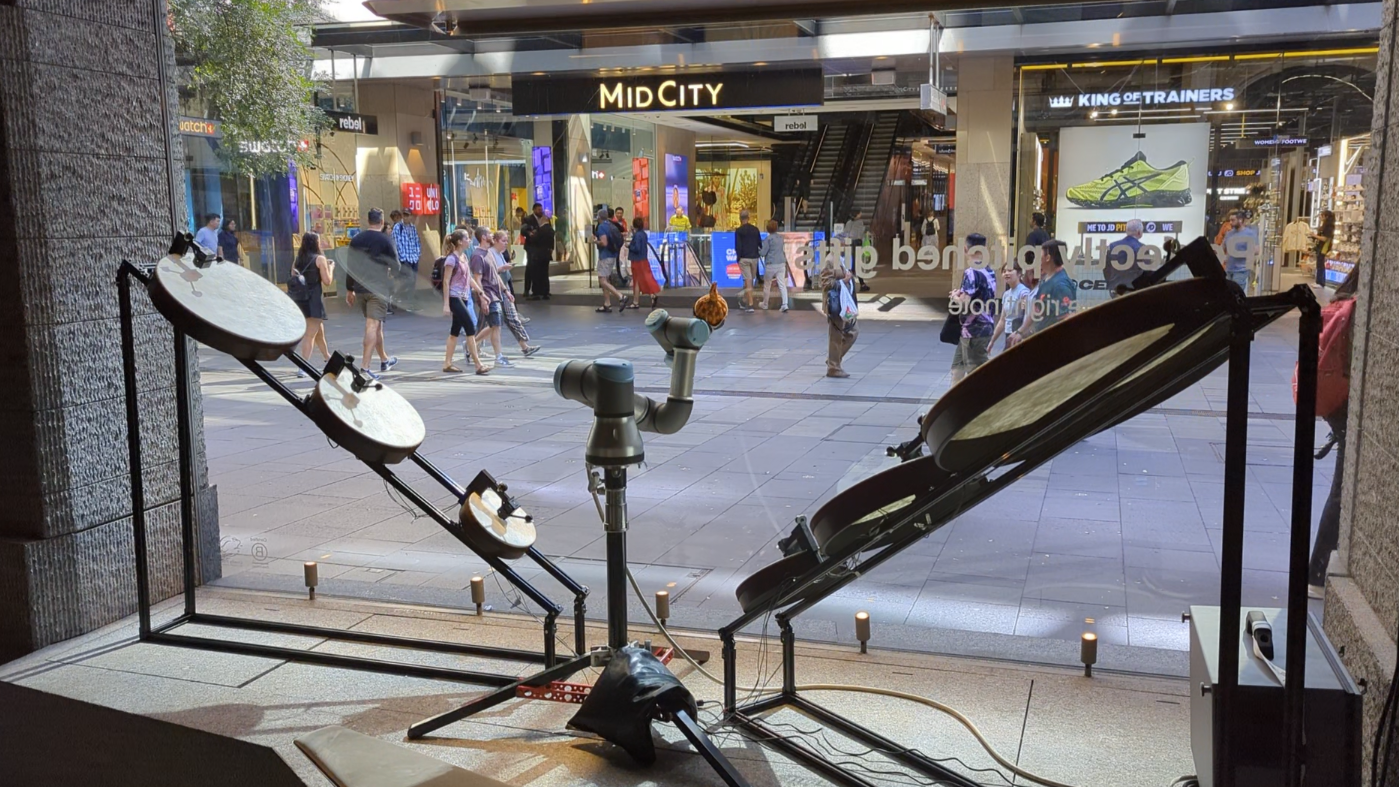}
	\caption{In-shop view of the installation}
	\label{fig:current.png}
\end{figure*}

\section{Introduction}\label{sec:introduction}
The integration of interactive music systems and robotic musicianship into commercial retail environment is a largely under-explored area~\cite{frid2023musical}. More commonly, such systems have been confined to laboratory settings or specific art installations and performances, where the controlled conditions differ vastly from the unpredictability and diversity of the real world \cite{sullivan2018stability}. Robotic musicians however have extensive potential in the real-world to engage with a range of audiences. 

This project centers around a three-week-long interactive in-store installation featuring a UR3 industrial cobot (collaborative robot) arm, custom-built frame drums, and an adaptive music generation system. The store front was in the city center of Sydney, Australia at AESOP, Pitt St. The installation was designed to interact with the ambient soundscape of the store, creating a dynamic and engaging musical experience for patrons. This paper demonstrates the technical feasibility of such an integration and also explores the artistic and commercial implications of embedding interactive robotic musicianship within a retail environment.

This project's core contributions include the development of a robotic arm integrated with actuated frame drums to forge a distinctive musical instrument, deploying interactive music to elevate the commercial store's ambiance, and showcasing the robotic system's resilience through its uninterrupted operation for a three-week span. Crucially, the project navigated the retail environment's unique challenges by prioritizing system reliability to avoid any disruptions or negative impacts on the customer experience. 



We also delve into the critical outcomes and reflections arising from the deployment of our interactive music system in a commercial context. Central to our discourse are the pivotal themes of system reliability, the nuanced dynamics of human-robot interaction within the retail sphere, and the implications of robotic agency as perceived by store staff and patrons. We finally explore the strategic design choices that underpinned the system's integration into the retail environment, ensuring its alignment with the store's brand identity while enhancing the customer experience. 

\section{Related Work}
\subsection{Robots, Music and Retail}
Recent studies have increasingly recognized the transformative potential of robots in retail environments, primarily focusing on their ability to enhance customer engagement and sales~\cite{de2020rise}. These include applications to enhance physical shopping experience, such as recommendations on purchases \cite{graef2023buy}. It is  unclear however, the extent to which robots can impact the shopping experience, with studies not pointing to clear findings about best practices to engage customers \cite{golchinfar2022let}. Some studies have used robots specifically for entertainment, such as SoftBank's Pepper Robot, to quiz customers about the store inventory \cite{de2021or}.  

Music plays a pivotal role in shaping the retail atmosphere, with many businesses curating licensed playlists to influence consumer behavior and enhance the shopping experience~\cite{michel2017thank}. Music has been shown to improve product evaluations and behavioural results in retail when effectively used, both through music choice and volume levels \cite{trompeta2022meta}. Like music, the strategic design of store-front displays is crucial in attracting customers and establishing a store's brand identity~\cite{cornelius2010storefront} and is extensively researched \cite{webber2018remodelling}.

\subsection{Robotic Musicianship}
Robotic Musicianship encompasses a multifaceted area of research that integrates musical mechatronics - concerning the mechanical systems that produce sound - and machine musicianship, which involves the development of algorithms and cognitive models for music perception, composition, performance, and theory \cite{savery2021shimonsing}. Drumming and percussion playing robots are likely the most widely addressed area of robotic musicianship. Projects have include egg-shaped rattles and drums controlled by direct actuation mechanisms \cite{singer2004lemur}. Many systems have been built that use solenoid motors for various percussion\cite{kapur2007comparison}. Recent approaches have focused on techniques such as using Brushless Direct Current (BLDC) motors for high speed and dynamic range \cite{yang2020mechatronics}. Other projects have focused on adding extra degrees of freedom, allowing for variation in where the drum is hit \cite{murphy2014little} or adding a flexible joint to the system to improve drum roll performance \cite{karbasi2022robotic}.

There have also been multiple projects extending findings from robotic musicianship into other fields. These include using approaches from robotic musicianship for real-time music generation to interact with individuals to complete basic tasks \cite{savery2021emotional, savery2021before}. Other applications include in education for children with autism \cite{taheri2019teaching} and custom devices for interaction \cite{savery2023enhancing}. This body of work, spanning from the development of percussion-playing robots to the application of these technologies in educational and therapeutic contexts, showcases the versatility and potential impact of robotic musicianship beyond traditional performance settings. 

\section{Frame Drums}\label{sec:introduction}
For this system, we chose to focus on the frame drums, selected for their versatility and adaptability to diverse musical styles, which aligns with the objective of creating an interactive music system that resonates with a broad audience. Their wide range of sizes and tunings, coupled with a potential uniform hitting mechanism, enabled the integration of these instruments into the robotic setup without necessitating significant alterations for each drum size.

\begin{figure}[t]
	\centering
		\includegraphics[width=\columnwidth]{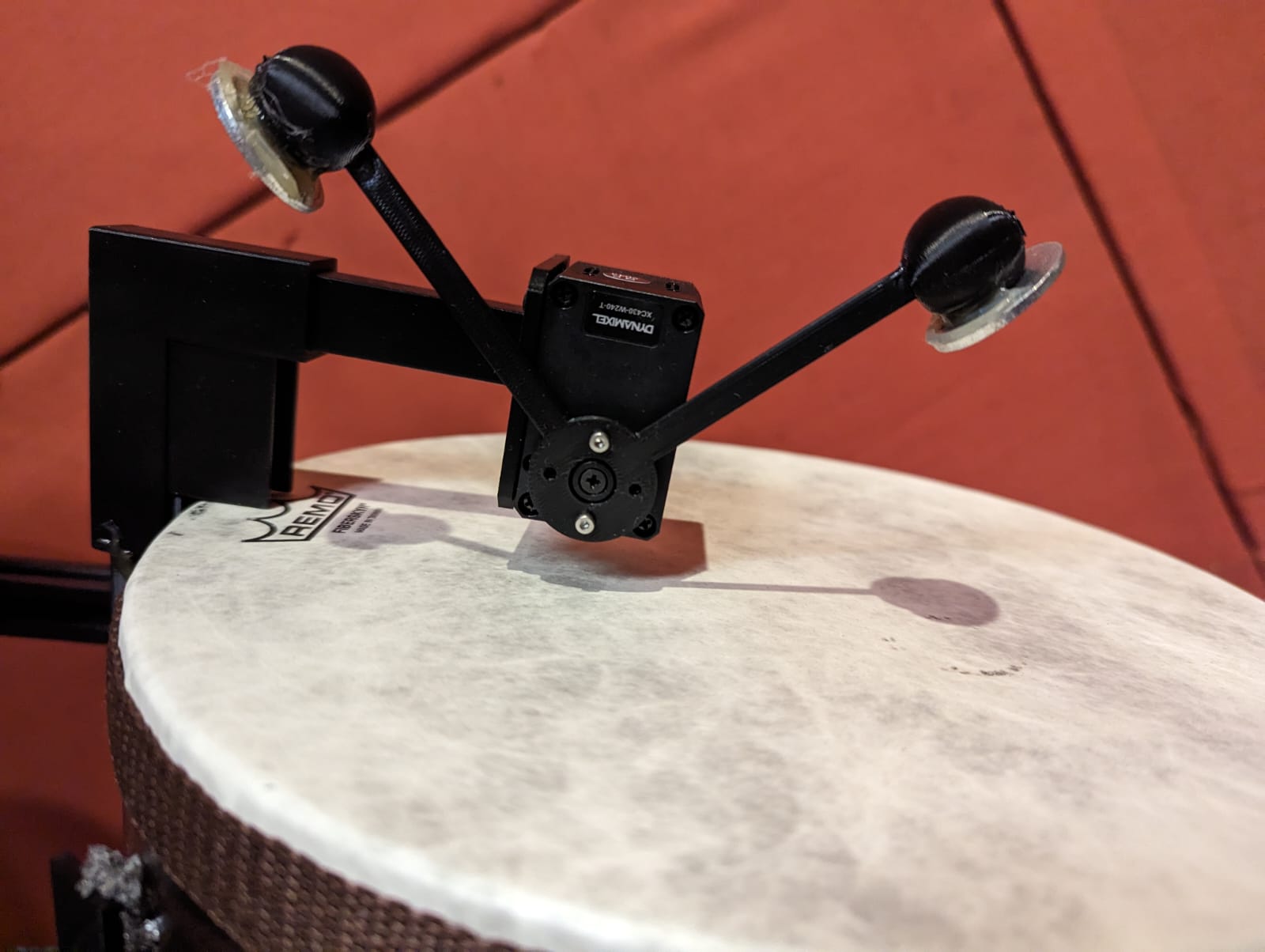}
	\caption{Close-up view of the drum actuation mechanism}
	\label{fig:beater}
\end{figure}

The physical design of the frame drum actuation system incorporates motors equipped with dual beaters, enabling hits on both the rim and the interior surface of the drum (see Figure \ref{fig:beater}). This design choice expands the tonal possibilities while also clearing displaying movement to the audience, through longer than required beater arms. Cabling for the actuation mechanism is routed through the beater holder, to maintain a clean appearance. The drums are positioned at a slant, primarily for aesthetic reasons, to draw attention to the central robotic arm and create a visually engaging setup that complements the store's ambiance.

The actuation mechanism is powered by six Dynamixel XC430 W240 T motors, controlled through a U2D2 motor controller. This setup facilitates precise control over the drum strikes, with each motor's range of motion pre-configured and stored within the motor's firmware. The entire system is managed via a Linux-based control platform, utilizing Python for motor control and system integration. This approach allows for real-time adjustments and synchronization with the central robotic arm, ensuring a cohesive performance.

The control software is designed to interface directly with the U2D2 motor controller, enabling dynamic control of the drum actuation system. Python serves as the backbone for the software architecture, providing a flexible and powerful tool for programming the musical sequences and integrating the frame drums with the broader system. This software layer is critical for achieving the desired musical output and ensuring that the frame drums contribute effectively to the interactive music experience.

The integration of frame drums into the robotic music system was designed around best practices for robotic musicianship, focusing on displaying movement to humans, even if at the cost of mechanical efficiency. This design choice reflects the overarching goal of the project: to create an engaging and immersive musical experience that leverages the unique capabilities of robotic systems while respecting the aesthetic and acoustic properties of traditional instruments.

\begin{figure}[]
	\centering
    \subfloat[UR3 robot arm with mounted caxixi.]{
		\includegraphics[width=0.9\columnwidth]{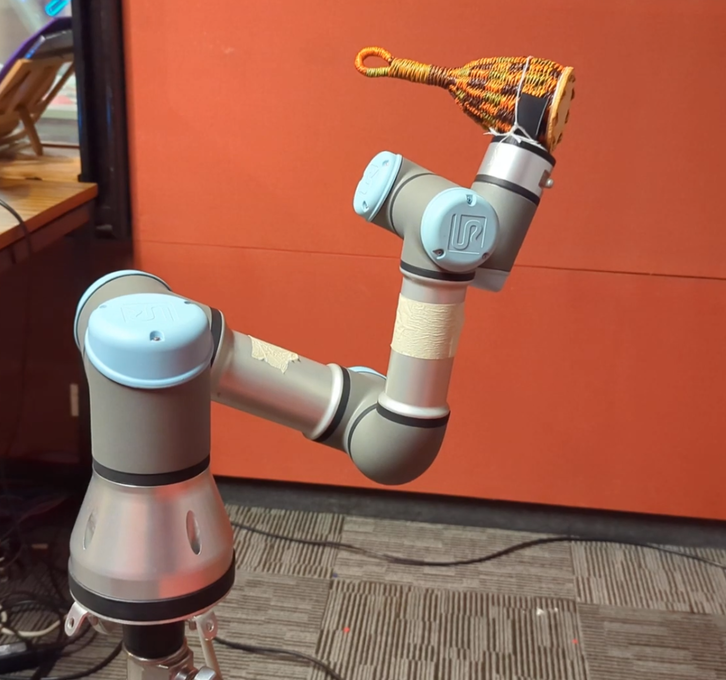}\label{fig:robot_sim}}\\
    \subfloat[UR3 robot arm with mounted caxixi.]{
		\includegraphics[width=0.9\columnwidth]{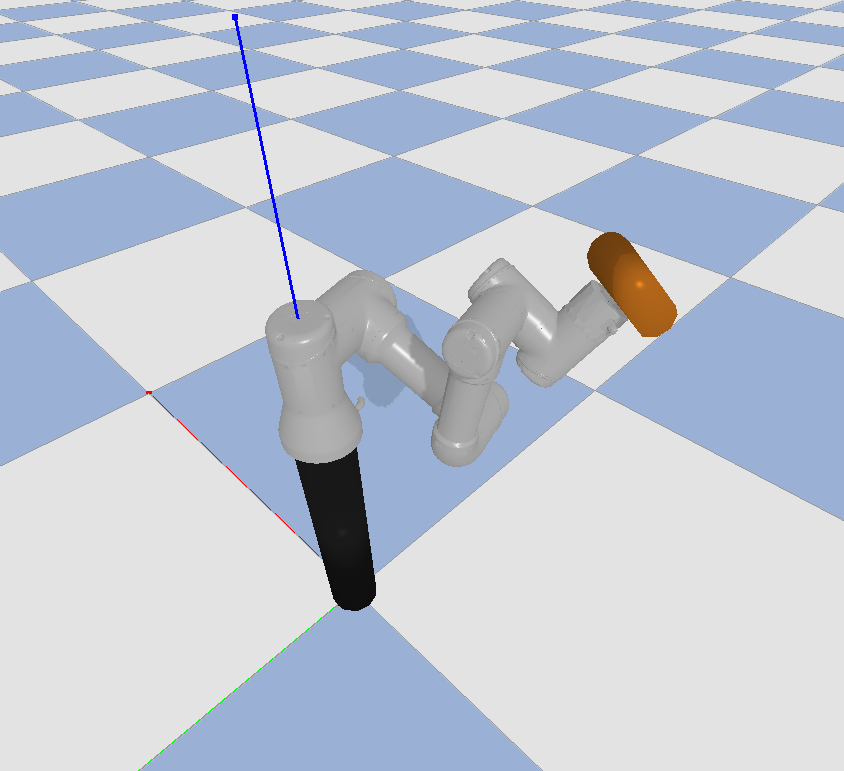}\label{fig:robot_hw}}
	\caption{Robot arm simulation and hardware setup.}
	\label{fig:hw_sim_setup}
\end{figure}

\begin{figure}[]
	\centering
		\includegraphics[width=0.9\columnwidth]{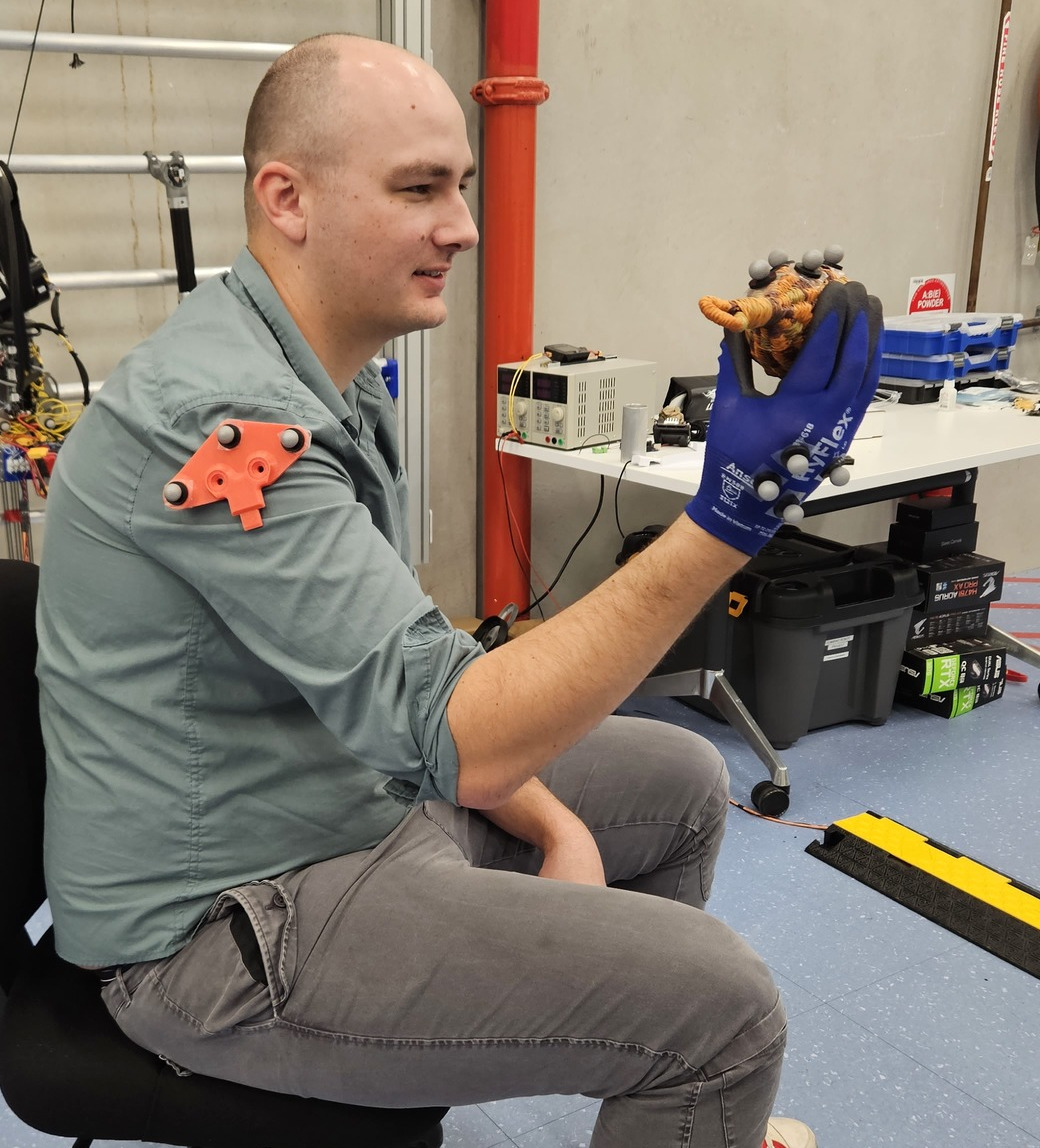}
	\caption{Motion capture system for recording human percussionist.}
	\label{fig:demonstration}
\end{figure}

\begin{figure*}[t]
	\centering
		\includegraphics[width=2\columnwidth]{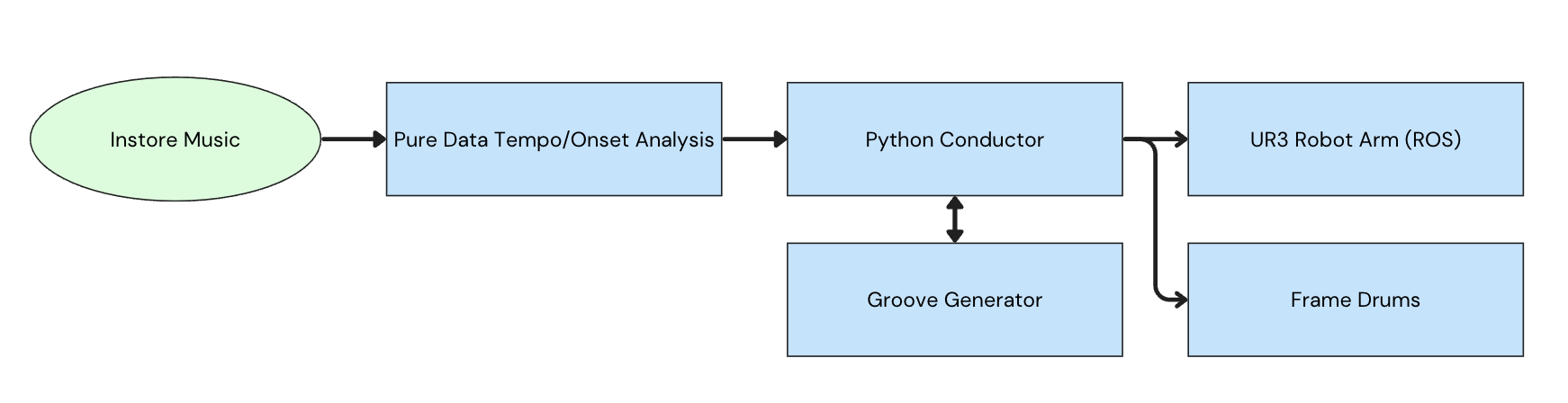}
	\caption{Flowchart of system}
	\label{fig:flowchart}
\end{figure*}
\section{Robotic Arm Percussionist}\label{sec:introduction}

For imitating a percussionist playing the caxixi, we utilised an off-the-shelf UR3 robot arm which is designed to be safe to operate around humans due to its auto-shutoff capability which utilises internal force sensing to detect unexpected external forces, such as colliding with an object. The UR3 is a highly capable robot with six degrees of freedom (DOF) which means that it can reproduce any arbitrary motion, within its workspace and motor limits, that a human can perform. Robotic arms have recently been used in multiple musical works, such as guitar performance and to convey dancing \cite{rogel2022robogroove}.

The caxixi was attached to the end effector of the robot using a 3D-printed mount, shown in Fig.~\ref{fig:robot_sim}. To execute motions on the arm we utilised Robotic Operating System (ROS)~\cite{ros} which has ready-to-use robot specific drivers. To simulate the robot and environment we used Pybullet~\cite{coumans2021} with OpenAI Gym and IKFast plugin~\cite{Cambel_ikfast}, shown in Fig.~\ref{fig:robot_hw}. This was used for ensuring the commanded trajectories were executable and also for avoiding environment collisions.

To program the playing of the caxixi we utilised a recording of a human percussionist using a VICON motion capture system. We track the trajectories of the both the percussionist's back palm and the caxixi relative to their shoulder. The marker setup is shown in Fig.~\ref{fig:demonstration}. The VICON captures 3D position and orientation, which we refer to as a pose, at 300 Hz.

To reproduce this trajectory on the robot arm we utilise Gaussian process (GP) regression, a non-parametric probabilistic interpolation method~\cite{Rasmussen2006}. First a GP is fitted to the trajectory and then using GP regression a smoothed out trajectory of poses at any arbitrary time sequence can be generated. The amount of smoothing can be controlled by adjusting the length-scale parameter of the GP kernel which is useful for ensuring the trajectory is executable on the robot arm which has physical joint velocity limits. More details can be found in~\cite{sukkar2023robotic}.

We then took three snippets from this recording with different rhythmic signatures and then translated this motion to the robot arm. For mapping this trajectory to the robot arm we calculate the inverse kinematics solutions for each pose. Since a 6-DOF robot arm can produce multiple solutions for the same pose, for example elbow-up or elbow-down, we bias these solutions to be visually close to the natural configuration of the percussionist's arm.
The robot arm also has an option of moving to a random pose to emulate a lively feel.
For ensuring executability, we experimentally chose a GP kernel length-scale which ensured the arm did not violate its joint limits or cause an unintended safety stop.




\section{Music Analysis and Generation}\label{sec:introduction}

For the music analysis and generation component of the installation, Pure Data (Pd) was selected for its ability to run on linux. The decision to utilize Pure Data (over pure Python) was also motivated by its capacity to provide a straightforward interface for store staff, ensuring that the system could be managed and troubleshooted with minimal technical expertise. This aspect was crucial for maintaining the system's operational integrity throughout its deployment in a commercial setting.

The Pure Data patch designed for this project was intentionally kept simple, focusing on stability and ease of use. The primary goal of the patch was to analyze the in-store music in real time and output two key pieces of information: the beats per minute (BPM) and the density of the music. This simplicity in design was essential for ensuring that the system could remain functional over extended periods without requiring complex interventions and avoiding (or at least lowering) the chance of crashing.

The analysis process begins with a volume threshold mechanism. If the in-store music's volume falls below this threshold, the system remains inactive, ensuring that the installation does not generate output in the absence of sufficient audio input. For detecting musical onsets, the patch employs the Bonk object, a widely used tool within Pure Data for onset detection. The number of detected onsets over time is then used to calculate the density of the music, providing a measure of its rhythmic complexity.

To determine the BPM of the in-store music, the patch utilizes the Beat~ object from the ELSE library by Alexandre Torres Porres. The patch monitors the stability of the detected BPM, and in cases of significant variation that suggests instability, it temporarily ceases to generate musical output. This safeguard ensures that the installation's responses are musically coherent and aligned with the ambient music.

After analyzing and detecting the BPM and density, a Python script creates drum rhythms. The generative systems uses a stochastic model that operates on the incoming level of density, combined with a random value for syncopation. Utilizing these parameters, the system allocates drum strikes across a two-bar grid, determining positions for drum hits. Elevated levels of syncopation facilitate an increased frequency of placements off the beat, whereas density levels dictate the quantity of notes generated. Subsequently, each note is given a probability of being played during each cycle. This secondary layer of probability ensures a continuous variability in the rhythm.

The design of the Pure Data patch was designed for flexibility to accommodate the diverse range of music played in the store, which included genres as varied as classical, jazz, pop, rock, and soundscapes. By focusing on the fundamental aspects of music analysis—tempo and density—the system was equipped to adapt to this wide array of musical styles. This adaptability was key to achieving the project's objective of enhancing the retail environment with an interactive musical experience that could dynamically respond to the store's soundtrack, regardless of its genre.

\section{System Integration}\label{sec:introduction}
The integration of the interactive music system is orchestrated through a central Linux laptop, which serves as the center for all operations. This centralized approach ensures cohesive control and synchronization over the system components. Audio input is analyzed by a Pure Data patch, which creates the density and BPM that are then streamed to a Python script. The script interprets the BPM and density data to generate a rhythmic grid for the frame drums, using a rule-based system to create varying patterns. This rhythmic grid, structured as a 2-bar pattern  repeats until there is a significant change in music density or when it has played for 32 bars, and dictates the activation signals for each frame drum hit and the robotic arm movement (See figure \ref{fig:flowchart}). The robot arm is sent either an instruction to play one of the three caxixi rhythms, or to move to a random pose in-time with the beat. The random pose ensures that the robot is relatively active, while not constantly playing the caxixi and distracting from the retail experience. 

One of the critical challenges in system integration is ensuring the synchronization of musical output with the detected beats. This requires considering any digital latency and accounting for the physical time required to actuate the frame drums and robotic arm. To manage this, the system implements a delay strategy for playtime, by the distance between each beats minus 120 ms, accounting for the movement time and system latency. This formula ensures that the actuation of the drums and robotic arm is precisely timed to match the rhythm of the music being analyzed.

To enhance the system's usability, especially for staff, auto-run scripts are configured on the desktop of the central Linux laptop. These scripts enable easy system start-up and restart, ensuring that the interactive music installation can be maintained with minimal effort and technical knowledge.

\section{Live Performance}
\begin{figure}[t]
	\centering
		\includegraphics[width=\columnwidth]{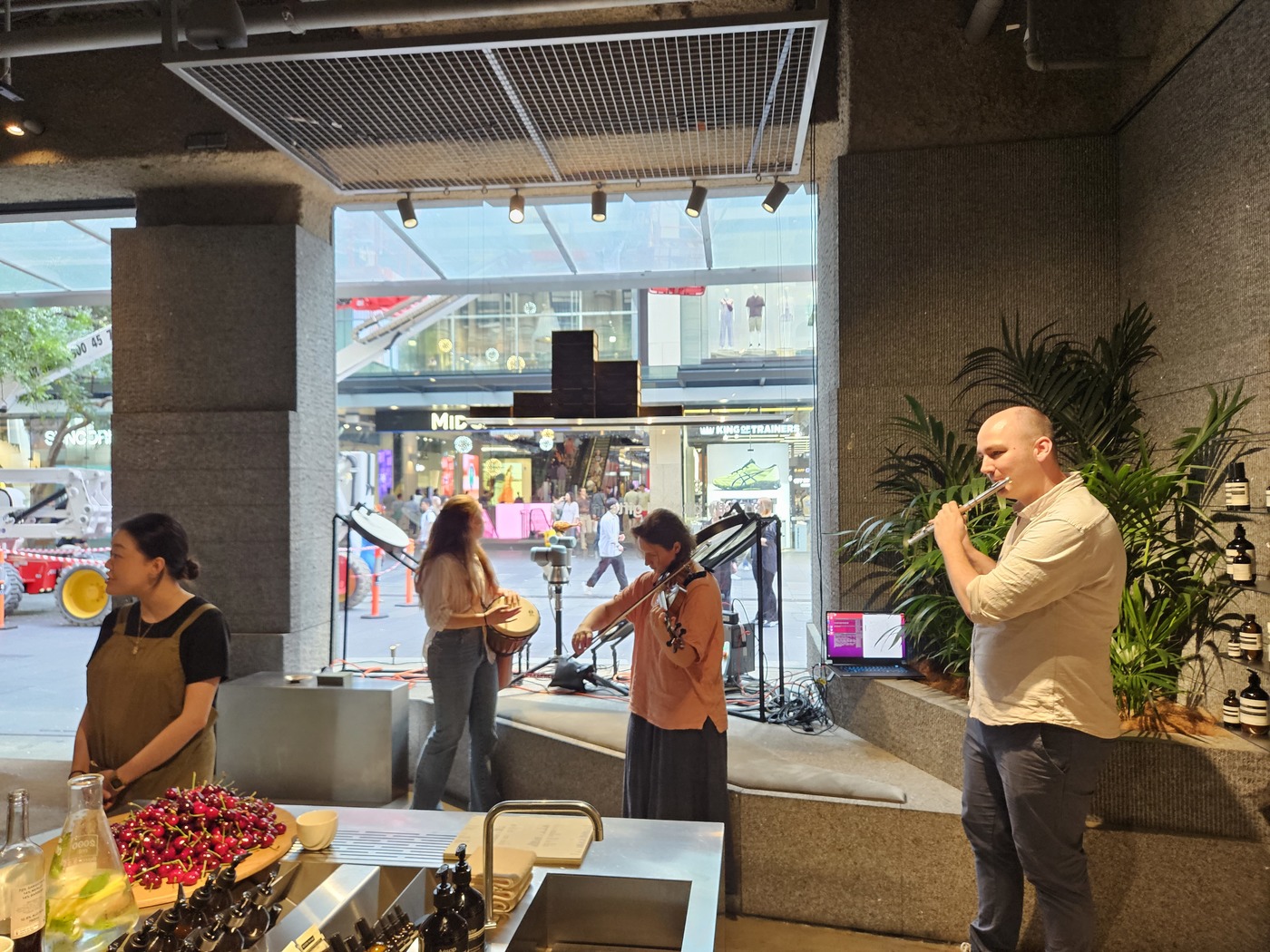}
	\caption{Human-robot live band performance}
	\label{fig:live_band}
\end{figure}

In conjunction with the installation, a special three hour live performance was organized towards the project's culmination (Fig.~\ref{fig:live_band}). The primary objective of this event was to explore the acoustic dynamics of the space further and to highlight the installation's presence and capabilities. This initiative served not only as a demonstration of the system's versatility but also as an opportunity to engage the audience directly with the interactive elements of the robotic musicianship in a more pronounced and focused setting. We invited musicians to improvise on violin, alto saxophone and djembe.

From a technical standpoint, modifications to the setup for the live performance were minimal. The key change involved the integration of a Rode Bluetooth clip-on microphone, which was employed to capture the audio from live musicians. This adjustment facilitated a direct and seamless connection between the musicians' performances and the installation, allowing the system to process and respond to live musical input in real time. The live performance was an exploratory venture into the soundscapes that could be generated within the given space, involving all combinations of the three instruments. This exploration was aimed at showcasing the dynamic interaction between human musicians and the robotic installation, drawing attention to the robot's role as an autonomous performer.

\section{Discussion}\label{sec:discussion}
This section reflects on the key outcomes and insights derived from the deployment of the interactive music system in a commercial storefront, emphasizing the significance of reliability, the perception of robotic agency, the dynamics of store operations, and the strategic design choices that influenced the system's integration and reception.

\subsection{Reliability is Key}
The operation of the interactive music system underscored the importance of reliability in public installations. Continuous operation without technical failures not only ensures the system's viability as a permanent fixture in a commercial environment but also builds trust and acceptance among the staff and patrons. We heard from staff that past installations in the space have been left not running after requiring constant intervention. Throughout the three weeks the frame drum continued to work, with the arm stopping only three times, likely due to a safety stop as a customer touched the arm. 

\subsection{Perceived Robotic Agency and Human-Robot Interaction (HRI)}
During the course of this project, a significant observation was made regarding the staff's interaction with the robotic elements within the retail environment. The staff's practice of anthropomorphizing the robotic system—evidenced by them assigning names to the robot and engaging with it on a daily basis—highlights a deep-seated human tendency to attribute human-like characteristics and agency to non-human entities. This behavior is not only indicative of the staff's acceptance and integration of the robot into their social space but also underscores the crucial role of anthropomorphism in fostering meaningful interactions between humans and robots, a phenomenon well-documented in the broader field of HRI \cite{blut2021understanding}.

We believe the inclination of the staff to engage in anthropomorphic behavior towards the robot was significantly influenced by the robot's interactive capabilities and its designated role within the retail environment. We noted on visiting that staff would note when the music stopped and discuss with Gary what the next song would be. Such interactions suggest that the perceived intelligence of the robot, along with its relational attributes—manifested through its capacity to musically engage and dynamically respond to the store's ambient soundscape—were instrumental in building a rapport and a sense of connection between the robotic system and human users. The anthropomorphic interactions not only facilitated a smoother integration of the robot into the daily routines of the store but also leveraged the unique capabilities of robotic musicianship to enrich the retail space, transforming it into an interactive and socially engaging environment.

\subsection{Approaches to Musical Generation}
The diversity of music that could potentially play in the store environment, ranging from classical to pop, jazz to electronic soundscapes, presented a significant challenge for designing a system that can appropriately respond to and interact with such a varied musical landscape. We initially considered curating custom playlists, however reconsidered in due to the importance of maintaining the existing store ambiance and branding. Altering the musical environment could inadvertently affect the overall store experience, potentially alienating customers who have specific expectations based on the store's established identity. Given these considerations, the approach to musical generation was focused on enhancing the interaction between the robotic system and the ambient soundscape of the store, and creating interesting movement and interaction.

\subsection{Evalutation in a Retail Setting}
Evaluating the impact of interactive music systems and robotic musicianship is challenging in any situation. The store we partnered with has a no-camera or recording policy for customer privacy, and directly asking customers was not an option. The effectiveness of such installations can instead be inferred from technical reliability, qualitative feedback from staff and customers, and the potential for future projects within the same venue. These indirect metrics serve as acted as our primary indicators of the system’s acceptance and performance, offering insight into its integration within the retail atmosphere and its influence on the customer experience.

In future installations, store Key Performance Indicators (KPIs) could play a role in assessing the system's success, focusing on metrics such as the increase in customers' desire to stay longer within the store and the extent of word-of-mouth sharing \cite{elmashhara2022linking}. An enhanced store atmosphere, attributed to the novel integration of robotic musicianship, is hypothesized to positively affect these KPIs. The assumption is that by creating a more engaging and interactive shopping environment, customers are more likely to enjoy their experience and share it with others, thereby amplifying the store's appeal and drawing in a wider audience. Despite the inherent difficulties in measuring the system's direct impact on specific customer behaviors, the positive feedback and technical performance offer substantial evidence of its value and effectiveness in enhancing the retail experience.

\subsection{Long-Term Effect and Novelty}
The concept of the novelty effect, as discussed by Reimann et al. (2023) in their study on a wine recommendation robot deployed in a supermarket, offers a pertinent lens through which to examine the interactive music system and robotic musicianship installed in the retail setting of our project \cite{reimann2023social}. In their research, Reimann observed that the novelty of the robot attracted individuals with no initial intention of buying wine, indicating that the robot's presence significantly influenced customer behavior and engagement patterns within the store. This divergence between expected and actual user interaction, coupled with the variance between behavioral data and survey responses, underscores the complexity of evaluating social robots in real-world settings and the impact of novelty on user engagement.

Drawing parallels to our work, the deployment of a robotic music system in a commercial environment likely elicited a similar novelty effect among patrons. The installation's uniqueness not only drew in the attention of shoppers but also potentially altered their in-store behavior, encouraging longer stays and increased interaction with the system. This phenomenon aligns with Reimann et al.'s findings, where the novelty of the robot became a focal point for customer interaction, irrespective of the initial intent to engage with the product being promoted.

Moreover, the observation by Reimann et al. that groups interacted more with the robot than individuals resonates with our experience, suggesting that the presence of innovative technology facilitates social interaction among store visitors. This behavior highlights the social dimension of technological novelty in retail spaces, where the introduction of interactive systems can transform shopping from a solitary activity into a collective experience.

A significant insight from Reimann et al.'s work is the importance of identifying and understanding the indicators of the novelty effect in field studies. By proposing a set of indicators and thresholds to recognize this effect, their research emphasizes the need to differentiate between short-term engagement driven by novelty and long-term interest based on genuine utility or enjoyment. Applying this perspective to our project, it becomes crucial to evaluate whether the interactive music system meets real user needs and continues to engage customers beyond its initial novelty.

\subsection{Dynamics of Store Operations and Feedback}
The support and involvement of the store staff were pivotal, highlighting the importance of designing systems that can be easily operated and managed by non-technical personnel. Feedback from the staff suggests a positive reception towards the technology, although it also emphasizes the need for systems that are robust and require minimal intervention to maintain their operation. This feedback is invaluable for future iterations, suggesting that while the novelty and engagement offered by such systems are well-received, their long-term success hinges on their ability to operate unobtrusively and reliably.

\subsection{Always On}
The decision to keep the system operational without requiring daily activation or deactivation by the staff presents an interesting case study in the deployment of interactive systems in public spaces. This approach minimized operational overhead for the staff and allowed for a continuous experience for patrons. However, it also raises questions about the long-term durability of the technology and the potential need for periodic maintenance or downtime to ensure continued optimal performance.

\subsection{Less Is More}
The strategic decision to design the system with high thresholds for beat consistency and tempo, prioritizing accuracy and relevance of musical output over frequency of interaction, reflects a nuanced understanding of the retail environment. In such settings, the potential distraction or annoyance of out-of-sync or overly frequent musical interjections outweighs the desire for constant engagement. This design philosophy underscores the importance of context in the deployment of interactive systems in retail, where the goal is to enhance rather than disrupt the ambient environment.

\section{Conclusion}\label{sec:introduction}
This project successfully demonstrated the feasibility and potential of integrating interactive robotic musicianship into a commercial retail environment. By deploying a UR3 arm alongside custom-built frame drums within a storefront, we showcased a novel application of automation in artistic expression but also enriched the ambient retail experience for customers. By focusing on enhancing the interaction between the system and the store's ambient music, rather than attempting to alter the store's musical landscape, we have demonstrated a novel approach to enriching the customer experience without compromising the store's brand identity.

Our exploration into the realms of anthropomorphism, human-robot interaction, and musical adaptability underscores the nuanced relationship between technology, music, and retail spaces. The spontaneous anthropomorphization of the robotic elements by store staff, attributing agency and character to the system, highlights the potential of such installations to become integrated, engaging parts of social and commercial settings. This project has also addressed the technical challenge of designing a system capable of responding to a diverse array of musical genres, emphasizing interaction and adaptability as central to the success of the installation. The system's ability to operate continuously for three weeks without significant interruptions highlights its reliability and suitability for long-term installations in public spaces. 

Moving forward, this project serves as a foundation for future research and development in the intersection of music, robotics, and retail. It invites further exploration into how these systems can be optimized to meet diverse user needs, adapt to changing environments, and contribute to the creation of innovative retail experiences. By continuing to focus on the seamless integration of technology with human-centric design principles, we can unlock new possibilities for enhancing public spaces with interactive, engaging, and meaningful installations.

\section{Acknowledgements}
This research is partially supported by the Industrial Transformation Training Centre (ITTC) for Collaborative Robotics in Advanced Manufacturing (also known as the Australian Cobotics Centre) funded by ARC (Project ID: IC200100001)

\bibliography{icmc2024template}

\end{document}